\documentclass[10pt,twocolumn,letterpaper]{article}

\usepackage{cvpr}              

\usepackage{graphicx}
\usepackage{amsmath}
\usepackage{amssymb}
\usepackage{booktabs}

\usepackage[pagebackref,breaklinks,colorlinks]{hyperref}

\usepackage[capitalize]{cleveref}
\crefname{section}{Sec.}{Secs.}
\Crefname{section}{Section}{Sections}
\Crefname{table}{Table}{Tables}
\crefname{table}{Tab.}{Tabs.}

\def\cvprPaperID{*****} 
\def\confName{CVPR}
\def\confYear{2023}

\begin{document}

\title{TempT: Temporal consistency for Test-time adaptation}

\author{Onur Cezmi Mutlu, Mohammadmahdi Honarmand, Saimourya Surabhi, Dennis P. Wall\\
Stanford University\\
{\tt\small \{cezmi, mhonar, mourya, dpwall\}@stanford.edu}
}
\maketitle


\begin{abstract}
     We introduce \textbf{Temp}oral consistency for \textbf{T}est-time adaptation (TempT), a novel method for test-time adaptation on videos through the use of temporal coherence of predictions across sequential frames as a self-supervision signal. TempT is an approach with broad potential applications in computer vision tasks, including facial expression recognition (FER) in videos. We evaluate TempT's performance on the AffWild2 dataset. Our approach focuses solely on the unimodal visual aspect of the data and utilizes a popular 2D CNN backbone, in contrast to larger sequential or attention-based models used in other approaches. Our preliminary experimental results demonstrate that TempT has competitive performance compared to the previous years' reported performances, and its efficacy provides a compelling proof-of-concept for its use in various real-world applications. 
\end{abstract}

\section{Introduction}
\label{sec:intro}
Affective computing aims to develop technologies with the capabilities like recognizing, interpreting, and simulating human affects. Expressions being one of the primary means of conveying emotion, facial expression recognition (FER) often constitutes an important part of human affective behavior analysis. There is an increasing number of use cases from driver safety applications to diagnosis and therapy of developmental problems of children \cite{kalantarian2019labeling}. With the continuous improvement in the computer vision field through extensive adoption of deep learning approaches, the real-world use of such algorithms is becoming easier and universal. However, the robustness and reliability of aforementioned algorithms tend to suffer from the domain shift phenomena which is still a prominent problem for computer vision models with limited generalization capability.

The domain shift problem becomes even more pronounced in the "real world" scenarios due to uncontrollable environmental conditions. In the computer vision setting some examples to these conditions could be lighting, camera quality, motion, and resolution. Invariance and robustness against these variations is the main focus of domain adaptation and domain generalization research, with many successful algorithms already developed. In our work, we explore a specific subdomain of this field called Test-Time Adaptation (TTA), also referred to as Unsupervised Source-Free Domain Adaptation. In this setting, we assume no access to the target domain during training-time and no access to target domain labels in test-time. We treat each video as a new domain and our method adapts the trained model to a given video during test-time to improve its performance.

\begin{table*}[ht]
  \centering
  \begin{tabular}{c|c|c|c|c|c|c|c|c|c}
    \toprule
    &Neutral&Anger&Disgust&Fear&Happiness&Sadness&Surprise&Other&\textbf{Total} \\
    \midrule
    Affwild2&44676&32962&7851&9730&2622&3296&5540&31412&\textbf{138089}\\
    Affectnet&55670&118605&19650&11647&5670&3626&19325&0&\textbf{234193}\\
    RAF-DB&3096&5771&2390&1571&347&865&846&0&\textbf{14886}\\
    \midrule   \textbf{Total}&\textbf{103442}&\textbf{157338}&\textbf{29891}&\textbf{22948}&\textbf{8639}&\textbf{7787}&\textbf{25711}&\textbf{31412}&\textbf{387168}
  \end{tabular}
  \caption{Label distribution of pretraining datasets}
  \label{tab:label_stats}
\end{table*}

We investigate the performance of our approach on the Facial Expression Recognition (FER) task, where the goal is to classify each frame in a video for Ekman emotions. The task of video assessment at the frame level is a natural environment for machine learning models with spatiotemporal inductive biases since the ability to model inter-frame relations could potentially be useful. Examples of such models are 3D convolutional neural networks (CNN) \cite{conv3d}, attention-based models \cite{transformer}, or hybrid approaches combining 2D CNNs with recurrent neural networks (RNN) \cite{cnnlstm}. The first two of these approaches usually suffer from greater computational requirements than 2D CNNs, whereas the last method has unstable training time behavior under inputs with longer duration. There are numerous solutions to these problems including more efficient architectures as well as well-studied training paradigms, but in our work, we focus on exploring an adaptive approach where a simple 2D CNN model, which lacks useful biases for the setting, uses temporal predictive consistency as a self-supervision signal to adapt at test-time. For benchmarking purposes, we use Affwild2 \cite{kollias2023abaw,kollias2022abaw,kollias2021distribution,kollias2021analysing,kollias2021affect,kollias2020analysing,kollias2019expression,kollias2019face,kollias2019deep,zafeiriou2017aff} which is  an invaluable FER dataset that contains over 500 videos and covers a wide variety of aforementioned variations. These qualities make it a suitable candidate for testing our algorithm.

\section{Related Work}
\noindent \textbf{Facial Expression Recognition (FER)}  is a challenging task, especially in real-world scenarios. The difficulty arises from the fact that there is a significant amount of variation within each expression category, making it difficult to distinguish between different expressions. Additionally, there can be similarities between different expression categories, which further complicates the task of FER.

This challenge is even more pronounced in real-world settings, where the lighting conditions, poses, and identities of the individuals can vary significantly. In such scenarios, even individuals with the same identity, pose, and lighting conditions can exhibit different expressions, while individuals with different identities, ages, gender, and pose can express the same emotion.

Thus, FER is a task that requires robust algorithms that can effectively handle these intraclass variances and inter-class similarities. In the past few years many Convolution Neural networks (CNN) based \cite{8453893, 8576656, 9075283} and transformer-based architectures \cite{xue2021transfer} have been proposed and significantly improve the performance of FER.

\begin{figure*}
  \centering
    \includegraphics[width=0.9\linewidth]{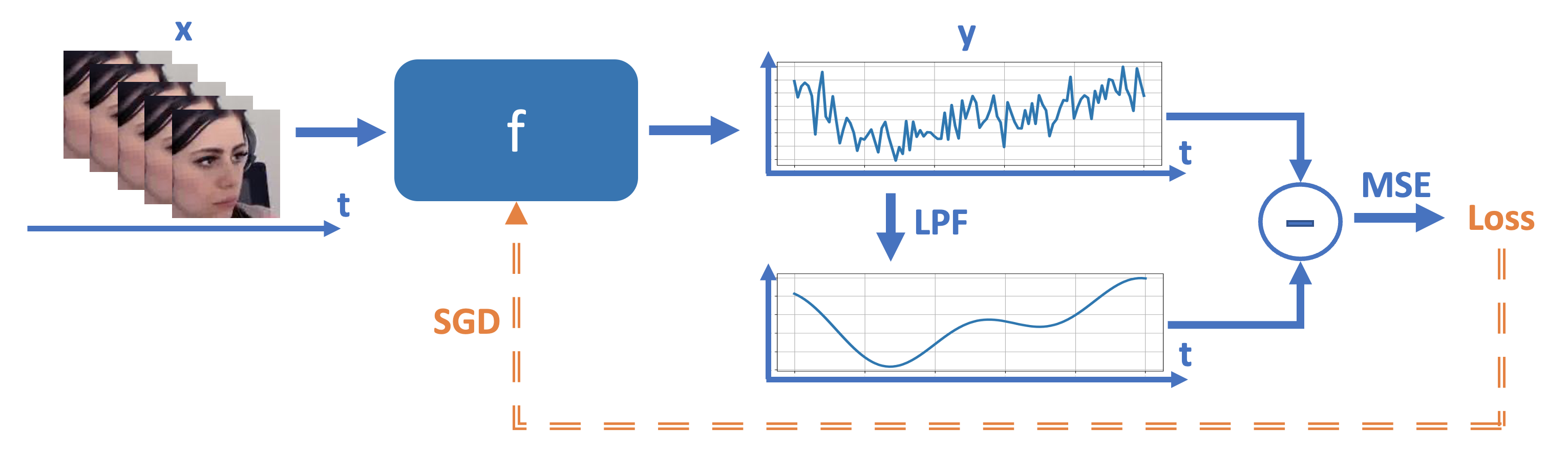}
    \caption{TempT algorithm}
    \label{fig:algo_figure}
\end{figure*}

As far as we are aware, there has been no prior research on test-time adaptation (TTA) for facial expression recognition (FER). Our work is an attempt to explore the use of TTA  on FER tasks. It represents a novel approach to FER that has the potential to improve accuracy and opens up new avenues for research into TTA.

\noindent \textbf{Test-time Adaptation} Early attempts for unsupervised domain adaptation were mainly based on updating running statistics of the batch normalization layers \cite{bn1, bn_adapt} with the new information from test data. \cite{ttt} was one of the early works to propose using an auxiliary self-supervised task to be used in the test-time with the purpose of adapting the backbone parameters. \cite{tent} proposed using entropy minimization as the main adaptation goal and limiting the set of parameters to be updated to the weights of batch normalization layers (as opposed to updating statistics as before) which are shown to be highly expressive in \cite{bn_expressive}. Originating from the close ties of domain adaptation with few-shot learning \cite{arm} introduces a meta-learning-based solution where the loss to be used for adaptation is meta-learned. Finally, \cite{memo} and \cite{shot} report impressive adaptation results by combining image augmentation and entropy minimization to overcome the shortcomings of the latter in scenarios with large domain shifts. 

All of these works operate on static data that does not necessarily bear temporal correlations. Among them, only \cite{tent} explores continual adaptation to online data streams. \cite{cotta} is a novel work that proposes a continual adaptation algorithm based on augmentation consistency. Yet, their algorithm makes an assumption of i.i.d. samples during test time, which may not always be correct. \cite{note} addresses this issue and coins a new normalization layer that handles selective adaptation under non-iid data streams. To our knowledge, none of the works in the field exploits the temporal correlations in a given stream, and in our work, we aim to explore a possible direction for that.

\section{Our Approach}

\subsection{Datasets and Preprocessing}

Focusing on training a computer vision model that operates on images (rather than videos), we have numerous data sources that are popular in the FER literature. We combine Affwild2 with Affectnet \cite{mollahosseini2017affectnet} 
and Real-world Affective Faces Database (RAF-DB) \cite{rafdb1, rafdb2} to create a larger and more diverse training dataset. In our task, target classes are 7 basic emotions (also known as Ekman emotions \cite{ekman}) plus an ``other'' class for expressions that do not fit into any category. 

\begin{figure}[!tbp]
  \centering
  \begin{minipage}[b]{0.4\textwidth}
    \includegraphics[width=\textwidth]{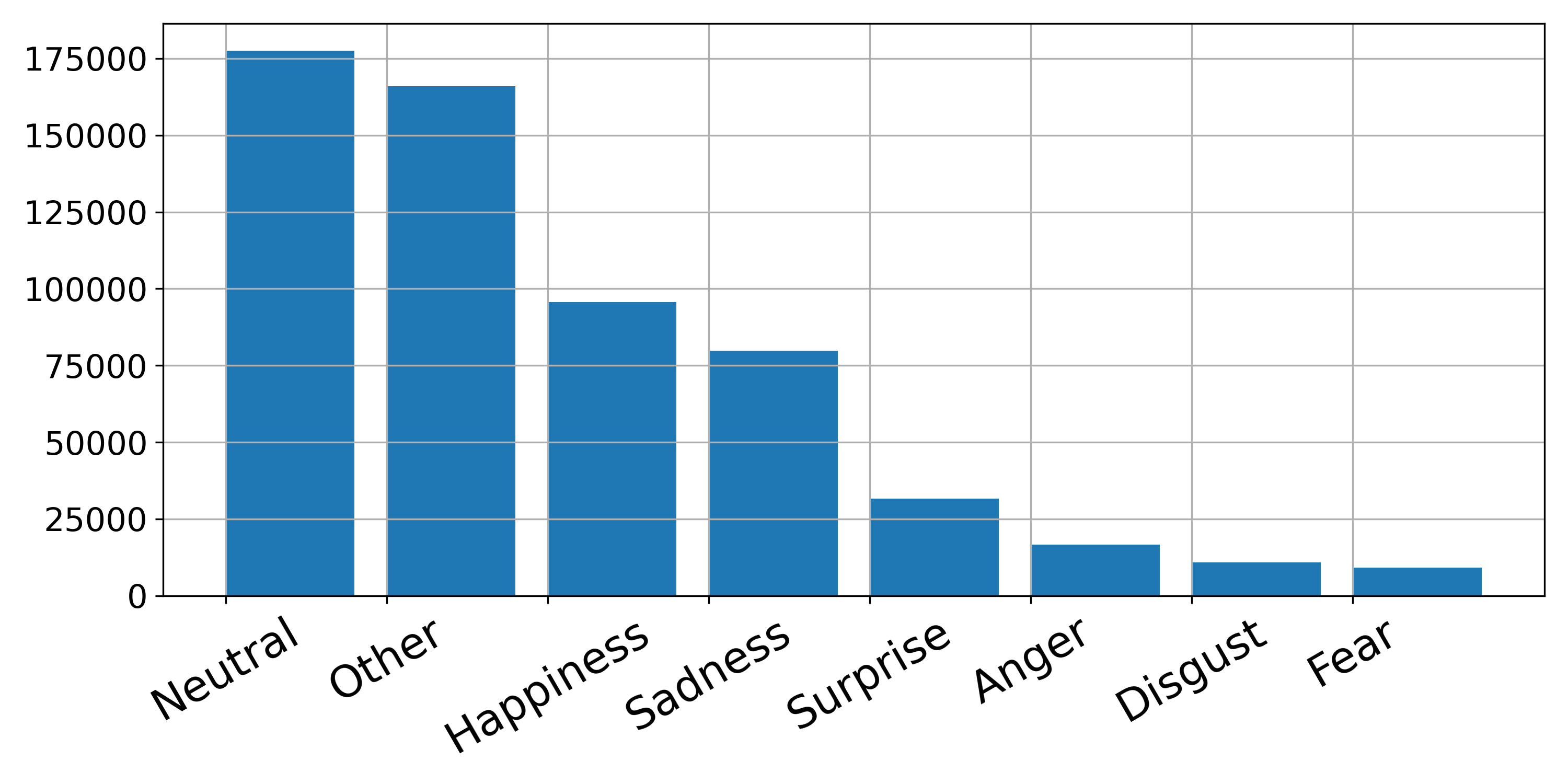}
    \caption{Label distribution of Affwild2 train set}
    \label{fig:affwild_train}
  \end{minipage}
  \hfill
  \begin{minipage}[b]{0.4\textwidth}
    \includegraphics[width=\textwidth]{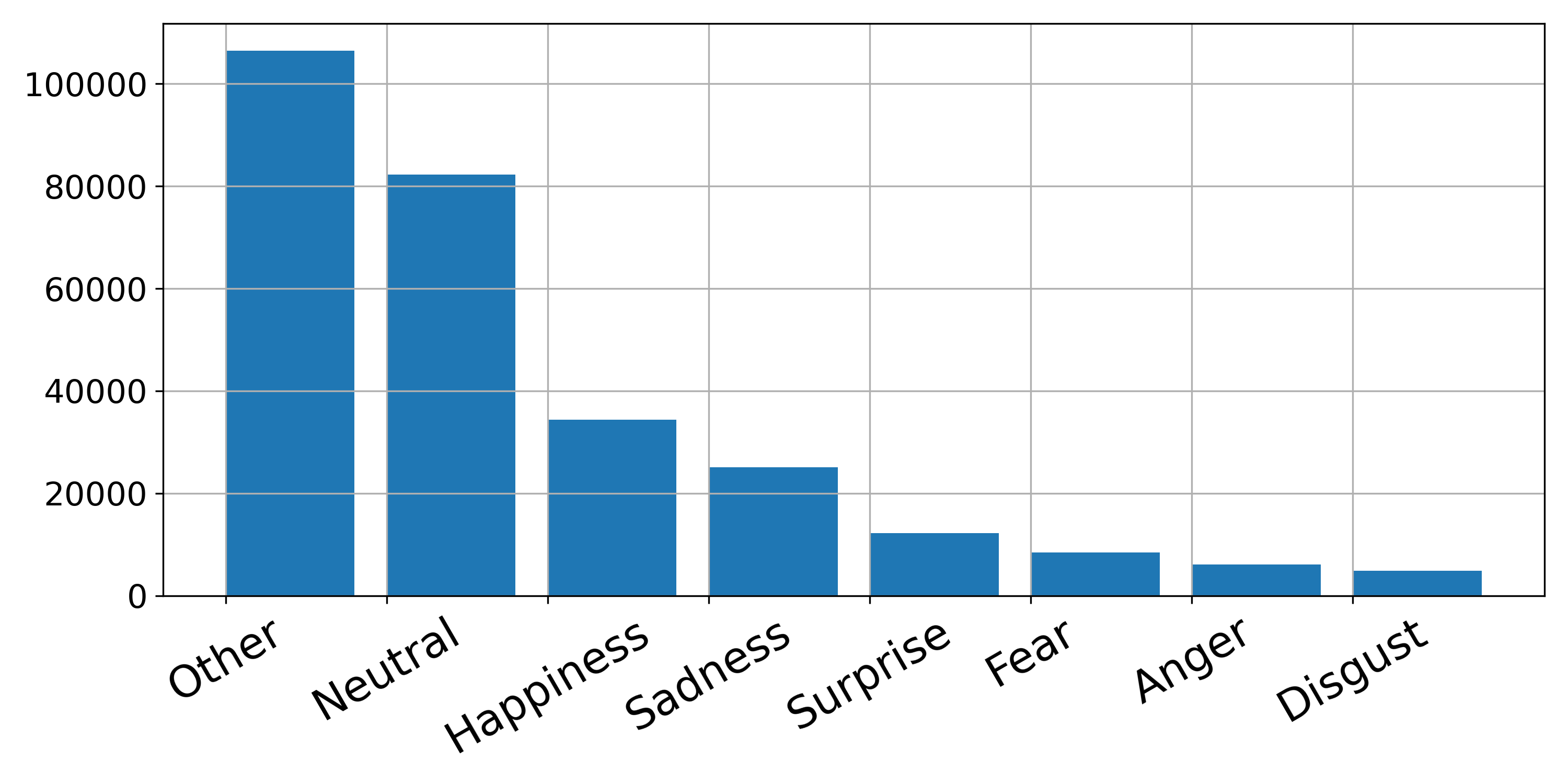}
    \caption{Label distribution of Affwild2 validation set }
    \label{fig:affwild_val}
  \end{minipage}
\end{figure}

Affwild2 is significantly larger in comparison to the others and has a label imbalance as can be seen in  \cref{fig:affwild_train} and \cref{fig:affwild_val}. In order to overcome this, we perform a random sampling on it by limiting the number of frames to 300  per video per expression class basis. Detailed label distribution of the resultant dataset is given in \cref{tab:label_stats}.

We use provided cropped and aligned images in Affwild2, and others are only available in cropped versions, so we do not require any additional spatial preprocessing for any of the datasets. We then resize images to $112\text{px}\times112\text{px}$ with antialiasing. For training purposes, we use common image augmentation methods such as color jitter, brightness and contrast shift, histogram equalization, channel dropout, blur, and random horizontal flip.

\subsection{Modeling}
Our approach is based on individual predictions on video frames, which allows us to use popular image-processing architectures in the literature. Due to their proven performance and stability of training, we use models from Resnet \cite{resnet} family, with variations such as aggregated residual transformations \cite{resnext} and squeeze-and-excitation blocks \cite{squeezeexcite}. Generated embeddings are processed by two fully-connected layers where the second, i.e. output, layer is subject to weight and input normalization \cite{salimans2016weight} to prevent overconfidence, improve smoothness, and generalization. 

Significant class imbalance is a problem in this setting that needs to be addressed for successful supervised training. Label weighting, class up-sampling, and class down-sampling are classic methods to alleviate this issue, yet there are numerous scenarios where they fail to do so. We, therefore, adopt another approach namely Label-Distribution-Aware Margin Loss (LDAM) that was introduced in \cite{ldam}. LDAM is similar to sample weighting in the sense that it modifies the loss depending on the class frequency but instead of using a multiplicative scaling, it intercepts with the class margins. The exact formulation is given in  \cref{ldam_loss} where $z$ is the unnormalized prediction vector,$y$ is the ground truth class label, $n_j$ is the number of samples in class $j$ and $C$ is a temperature-like hyperparameter that tunes the effect of margins. LDAM enforces larger margins on minority classes which in return increases the model robustness and prevents overfitting. For more details, we refer the reader to the original paper.

\begin{equation}
\begin{split}
\label{ldam_loss}
    \mathcal{L}(z,y) &= - \log\frac{e^{z_y-\Delta_y}}{e^{z_y-\Delta_y} + \sum_{j\neq y}e^{z_j}} \\
    \text{where  } \Delta_j &= \frac{C}{n_j^{1/4}} \text{ for } j\in \{1, \dots, k \}
\end{split}
\end{equation}

Supervised training of the model is then performed with back-propagation algorithm using defined LDAM loss  to account for the skewed label distribution. Adam \cite{adam} optimizer with weight decay \cite{adamw} is used for optimization where learning rates were subject to a step-decay schedule. Modeling and training were performed using PyTorch \cite{paszke2019pytorch} framework on NVIDIA V100 GPUs.

\subsection{TempT: Temporal consistency for Test-time adaptation}
\begin{figure*}
  \centering
  \begin{subfigure}{\linewidth}
  \centering
    \includegraphics[width=0.9\linewidth]{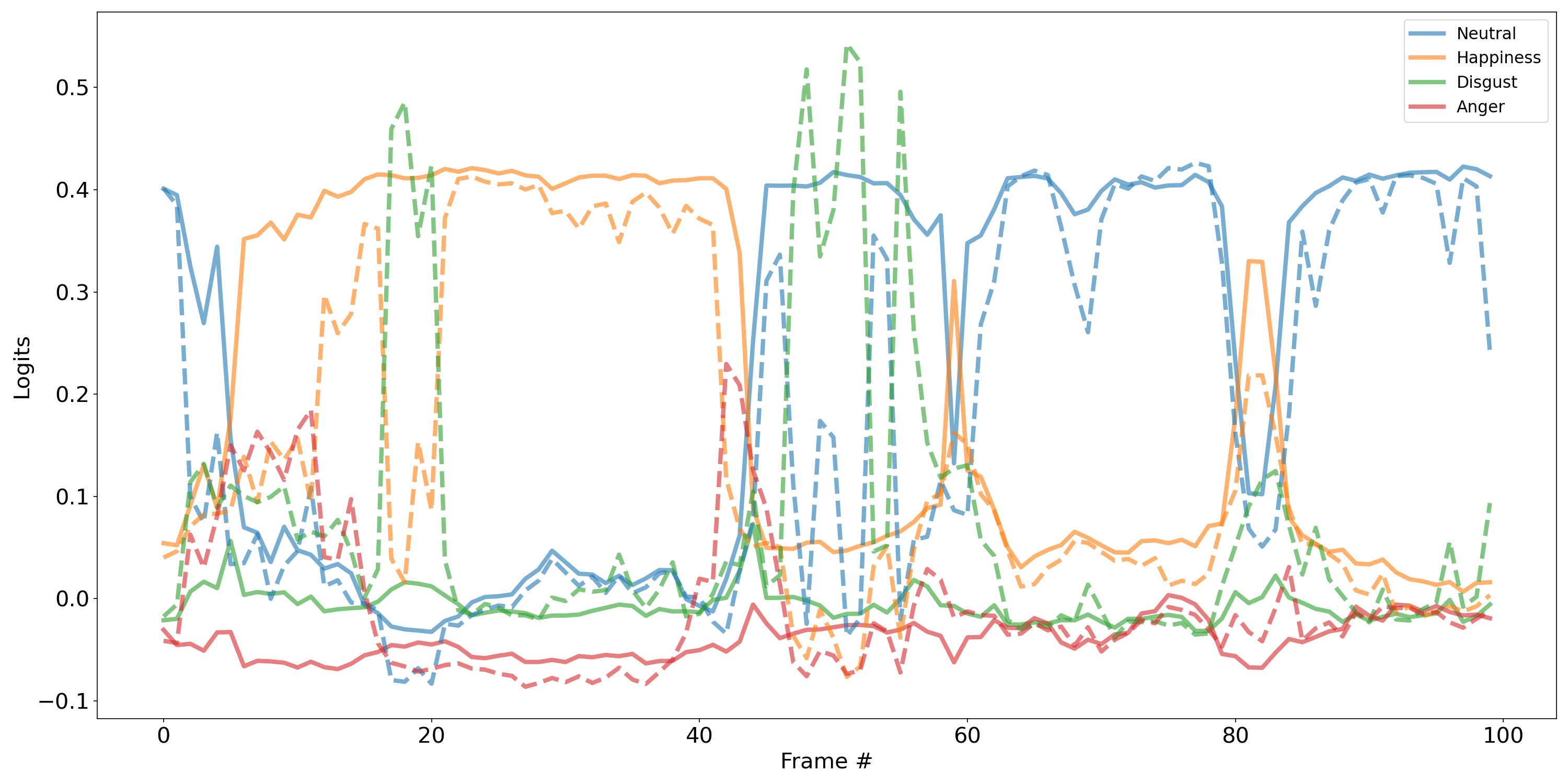}
  \end{subfigure}
  \begin{subfigure}{\linewidth}
  \centering
    \includegraphics[width=0.9\linewidth]{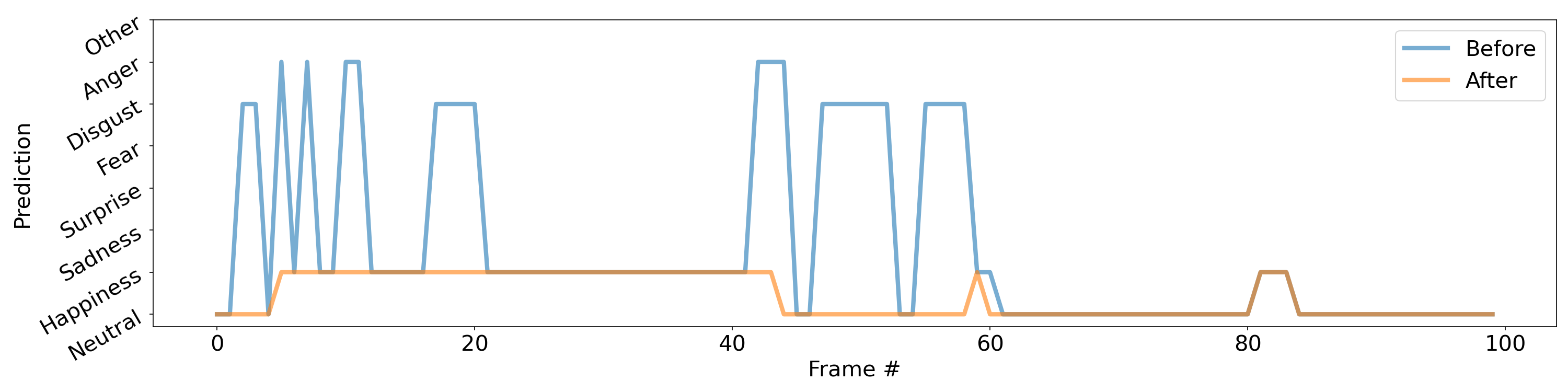}
  \end{subfigure}
  \caption{(top) Model outputs before (dashed) and after (solid) adaptation. (bottom) Model predictions before and after adaptation}
  \label{fig:before_after}
\end{figure*}

Being trained on static images as opposed to videos, 2D CNN models do not carry the implicit bias for the smoothness and/or consistency in their predictions across frames. We found empirically that such models contain stronger high-frequency components at the output, and when they are subject to a low-pass filter, the results look more desirable. We propose using this fact to generate a supervision signal to tune the network and improve classification performance. In particular, we temporally smooth the model predictions using a low-pass filter and set it as the desired signal. Purpose of setting this filtered signal as the target is to enforce the model to make temporally consistent predictions. We then calculate the mean-squared error between the original and target signals and use back-propagation to update a subset of model parameters.  

More formally, let $x^{(t)}\in \mathbb{R}^{112\times112\times3}$ be the $t^{th}$ frame of video and $f(.):\mathbb{R}^{112\times112\times3}\rightarrow\mathbb{R}^{8}$ be the trained neural network of interest. We hypothesize that predictive coherence between consecutive samples can be used as an implicit Jacobian regularizer. In \cite{jacobian}, it has been shown that regularization on the Frobenius norm of input-output Jacobian  of a neural network can help the network attain flatter minima with higher robustness against input variations. Now, consider the case when the frame rate of a video is high enough. We can then approximate the Jacobian as in \cref{jacobian}. 

\begin{equation}
\label{jacobian}
    J_{i,j}(x^{(t)}) = \frac{\partial f_i(x^{(t)})}{\partial x_j^{(t)}} \approx \frac{f_i(x^{(t)})-f_i(x^{(t-1)})}{x_j^{(t)}-x_j^{(t-1)}}
\end{equation}

Then minimizing the Frobenius norm of the Jacobian becomes equivalent to minimizing inter-frame prediction differences as in \cref{frob_minimize}

\begin{equation}
\begin{split}
\label{frob_minimize}
    \min\|J(x^{(t)})\|_F &\equiv \min\sum_{i,j} J^2_{i,j}(x^{(t)}) \\ &\equiv \min \|f_i(x^{(t)})-f_i(x^{(t-1)})\|
\end{split}    
\end{equation}

We empirically found that the initial distribution of prediction differences is heavy-tailed, with the tail being caused by momentary jumps in predictions due to problems at input cropping and/or sharp changes in activations due to model imperfections. When we used the target in \cref{frob_minimize} these outliers made the training process unstable for a significant portion of the experiments. We, therefore, chose to use another equivalent formulation to minimize the target. We first pass all frames from the pipeline to obtain an initial set of unnormalized scores $y^{(t)}\in\mathbb{R}^{8}$. We then use the error signal in \cref{eq:loss_signal} as a self-supervision loss function to fine-tune the model. $LPF(.)$ can be any low pass filter; in our experiments, we use a median filter, due to its robustness to outliers.

\begin{equation}
\label{eq:loss_signal}
    \mathcal{L}(y) = \sum_{t} \| y^{(t)} - LPF(y)^{(t)} \|
\end{equation}

Using the entire video for adaptation may not be computationally feasible when the video duration is long. To alleviate this, we count the number of changes in model predictions using a sliding window and select the regions with the most changes to be the training regions that will compose the training batch. The updated version of the loss signal can be examined in \cref{eq:loss_signal_batch} where $\mathcal{R}$ is the set of selected regions, and $r$ indicated the range of frames to be considered.

\begin{equation}
\label{eq:loss_signal_batch}
    \mathcal{L}(y) = \sum_{r \in \mathcal{R}} \sum_{t \in r} \| y^{(t)} - LPF(y)^{(t)} \|
\end{equation}

Being differentiable, this loss allows the use of backpropagation to update model parameters. The choice of parameters has an important effect on the performance of the adapted model since the selection defines the expressivity of the model and therefore the power of adaptive interventions. Following the analysis in \cite{bn_freeze_adapt}, we select this subset to be the weight and bias terms in batch normalization layers while freezing the running statistics. This has been shown to yield enough expressivity while preventing overfitting. We then use AdamW optimizer with learning rate set to 0.0001, for the adaptation process and take 10 gradient steps, a number that has proven empirically optimal in our hyperparameter searches. 

\section{Experiments}

We test TempT on the AffWild2 dataset and compare the results against the baseline model as well as another test time domain adaptation method, namely TENT \cite{tent}. We performed an extensive hyperparameter search on the adaptation parameters of TempT, such as the number of steps, learning rate, optimizer, etc., and report the performance of the best configuration in \cref{tab:results}. Static models' performances are deterministic whereas for adaptation cases we report an average F1 score over 20 experiments to account for stochasticity arising from a random sampling of adaptation frames. We clearly see the positive effect of TempT on classification performance. One important observation of these results is the ability of adaptation to help a less complex model reach the performance of a much larger one. In this experiment, SE-ResNext-101 has 8 times the number of parameters of Resnet-18. Another observation of the results is the performance disruption that TENT introduces. It consistently hurt the performance of the baseline model and we argue that this is due to the highly correlated inputs that we have during test time, which is predicted in \cite{note}. 

To further observe the changes that TempT induces, we also investigate time series generated by the model before and after adaptation. In \cref{fig:before_after} we provide such an example taken from 100 frame portion of a validation set video. On the top figure we provide unnormalized model outputs before and after adaptation, whereas bottom figure shows 'argmax' predictions. To create a cleaner top plot we omitted the classes that do not become the dominant prediction during this interval. From this visualization we can see, in a qualitative manner, that adaptation reduces the flickering behavior at the output and provides more coherent predictions over time, while increasing the F1 score for this particular video from 0.39 to 0.47. To have a quantitative understanding of this effect, we computed average number of decision changes before and after the adaptation on entire AffWild validation dataset. TempT reduces normalized number of changes (i.e. number of changes per frame) for a given video from 0.15 to 0.043. 

\begin{table}
  \centering
  \begin{tabular}{c|c|c|c}
    \toprule
      & Supervised & TENT\cite{tent} & TempT \\
    \midrule
    Resnet-18 & 0.307 & 0.277 & 0.323\\
    SE-ResNext-101 & 0.325 & 0.269 & 0.345 \\
    \bottomrule
  \end{tabular}
  \caption{Average F1 Score performances on validation set}
  \label{tab:results}
\end{table}




\section{Conclusion and Future Work}
In our work, we explored a novel model-agnostic algorithm that can have real-life applications for similar tasks and showed that this adaptive method could enhance model performance without any additional means of supervision. On the other hand, performance variance due to stochasticity in the frame sampling process is a problem that needs to be addressed to obtain a more deterministic understanding of the limits and behavior of the algorithm. With such increased stability and the performance boost it brings, TempT can potentially enable more reliable use of models on edge devices while protecting user privacy.

\section{Acknowledgements}
We would like to thank all members of the Wall Lab for providing valuable feedback. The work was supported in part by funds to DPW from the National Institutes of Health (1R01LM013364-01, 1R01LM013083), the National Science Foundation (Award 2014232), Lucile Packard Foundation (Auxiliaries Endowment) the ISDB Transform Fund, and program grants from Stanford's Human-Centered Artificial Intelligence Program, and from the Wu Tsai Neurosciences Institute's Neuroscience:Translate Program.

{\small
\bibliographystyle{ieee_fullname}
\bibliography{egbib}
}

\end{document}